\documentclass{article}
\usepackage{spconf,amsmath,graphicx}
\usepackage{hhline}
\usepackage{bm}
\usepackage{bbm} 
\usepackage{tabularx}
\usepackage{multirow}
\usepackage{booktabs} 
\usepackage{mathrsfs}
\usepackage{tabulary}
\usepackage{makecell}
\usepackage{amsfonts,amssymb}
\usepackage[ruled,linesnumbered]{algorithm2e}  
\usepackage{tabularx}
\usepackage{color, colortbl}
\definecolor{Gray}{gray}{0.8}
\definecolor{LightGray}{gray}{0.8}

\title{Improving Long Text Understanding with Knowledge \\ Distilled from Summarization Model}
\name{Yan Liu$^1$, Yazheng Yang$^2$, Xiaokang Chen$^3$}
\address{
$^1$The Chinese University of Hong Kong,\\ 
$^2$Department of Computer Science, Hong Kong University,\\
$^3$School of Intelligence Science and Technology, Peking University,
}

\begin{document}
%
\maketitle
\begin{abstract}
Long text understanding is important yet challenging for natural language processing. A long article or document usually contains many redundant words that are not pertinent to its gist and sometimes can be regarded as noise. 
With recent advances of abstractive summarization, we propose our \emph{Gist Detector} to leverage the gist detection ability of a summarization model and integrate the extracted gist into downstream models to enhance their long text understanding ability.
Specifically, Gist Detector first learns the gist detection knowledge distilled from a summarization model, and then produces gist-aware representations to augment downstream models.
We evaluate our method on three different tasks: long document classification, distantly supervised open-domain question answering, and non-parallel text style transfer. The experimental results show that our method can significantly improve the performance of baseline models on all tasks.
\end{abstract}
\section{Introduction}
\label{sec:intro}
\begin{figure}[t]
\centering
\includegraphics[width=\linewidth]{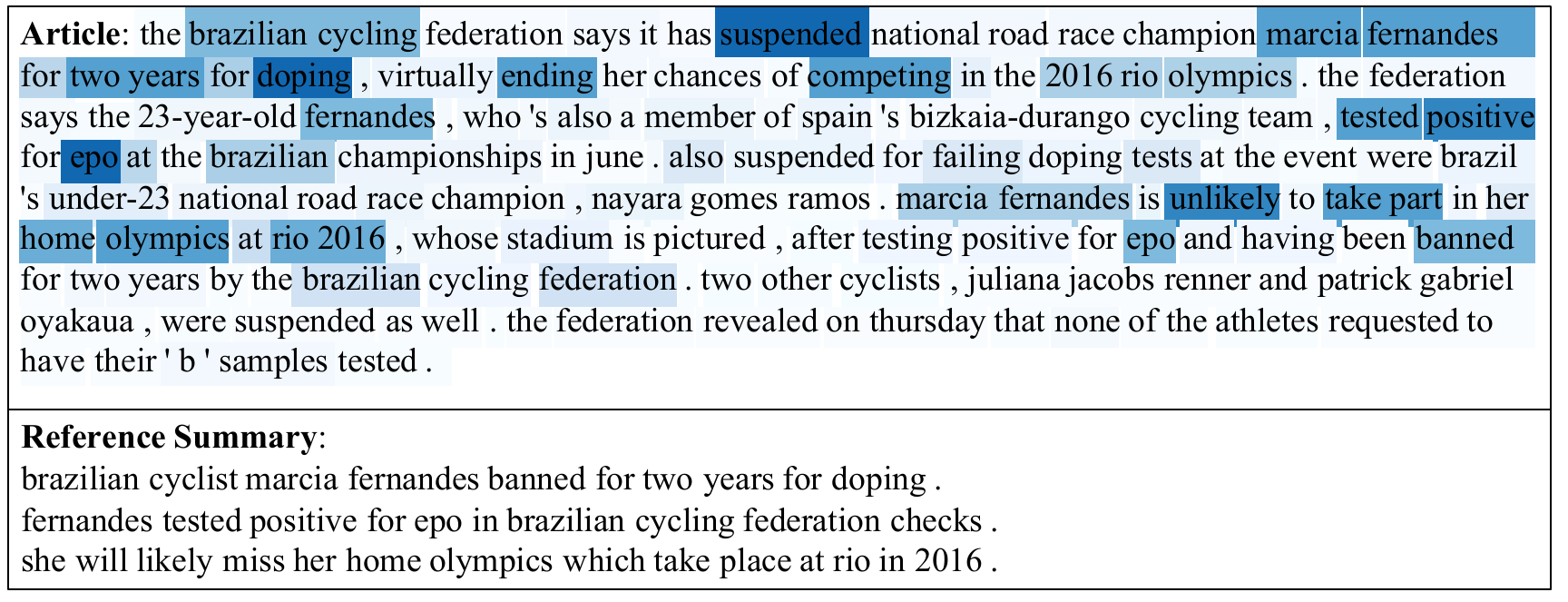}
\caption{\label{fig1} An example from the CNN/Daily Mail dataset. The shading intensity represents the importance weight extracted from a well-trained summarization model.}
\vskip -0.1in
\end{figure}

Recently, deep learning has developed rapidly \cite{chen2022context,chen2022d,chen2022group,tang2022not,chen2022conditional,meng2021conditional,chen2021semi,chen2020bi,chen20203d,chen2020real,tang2022compressible,tang2022point}. Transformer-based models are prevalent~\cite{liu2023parallel,liu2023uncovering,liu2023uncovering2,zan-etal-2022-language,zan2023programming,zan2023large} across numerous NLP tasks\cite{liu2022mpii}, but have difficulty in processing long texts due to the quadratic complexity of input text length\cite{eltu2022}.
Unlike short texts, long texts intrinsically contain many noisy words irrelevant to their gist. 
Although recent works have achieved promising results, few of them pay attention to measuring whether each part of the text is salient or negligible.
Abstractive summarization is a classic NLP task which aims to compress and rewrite a source text into a short version while retaining its main information \cite{DBLP:journals/corr/abs-2103-11332,DBLP:journals/corr/abs-2102-09130}. 
With this optimization objective, a well-trained summarization model has the potential to detect the gist of long texts.
Figure \ref{fig1} shows an example from the \emph{CNN/Daily Mail} \cite{nallapati2016abstractive} dataset, where the blue shading intensity represents the importance weight extracted from a well-trained summarization model. As we can see, the summarization model learns to focus on gist-relevant parts while neglecting irrelevant ones.
Intuitively, the gist detection ability can improve long text understanding through making models aware of salient parts of long texts.

\raggedbottom

In this paper, we propose to leverage the gist detection ability of a summarization model and integrate the distilled gist information into downstream models to enhance their long text understanding ability. 
However, there remain two challenges:
First, it is time-consuming to extract salient information from a large summarization model for each training sample.
Second, the summarization model produces salient information at each decoding step, while long text understanding models produce a single representation.

To solve these challenges, we propose our Gist Detector to transfer the gist information from a summarization model to downstream long text understanding models.
Specifically, Gist Detector is first trained to reproduce the gist information from the summarization model, then provides the gist-aware representation as supplementary to augment long text understanding models.
We train our Gist Detector with knowledge distillation mechanism, where a summarization model with an encoder-decoder architecture is the teacher model and Gist Detector with a fewer-layers' encoder is the student model.
The student model is trained with the average attention distribution over all decoding steps produced by the teacher model as "soft target".
Since Gist Detector is a non-autogressive model and much smaller than the summarization model, the process of gist extraction can be significantly efficient.
Then, we integrate the gist information extracted by our distilled Gist Detector into downstream models with a fuse module, effectively enhancing their long text understanding ability.

To evaluate the effectiveness of our method, we conduct extensive experiments on three tasks:  long document classification, distantly supervised open-domain question answering (DS-QA) and non-parallel text style transfer. Experimental results reveal that our method effectively augments different baseline models with better long text understanding ability, thus achieving significant performance improvement on all downstream tasks.

\begin{figure}[t]
\vskip -0.1in
	\begin{center}
		\includegraphics[width=1\linewidth]{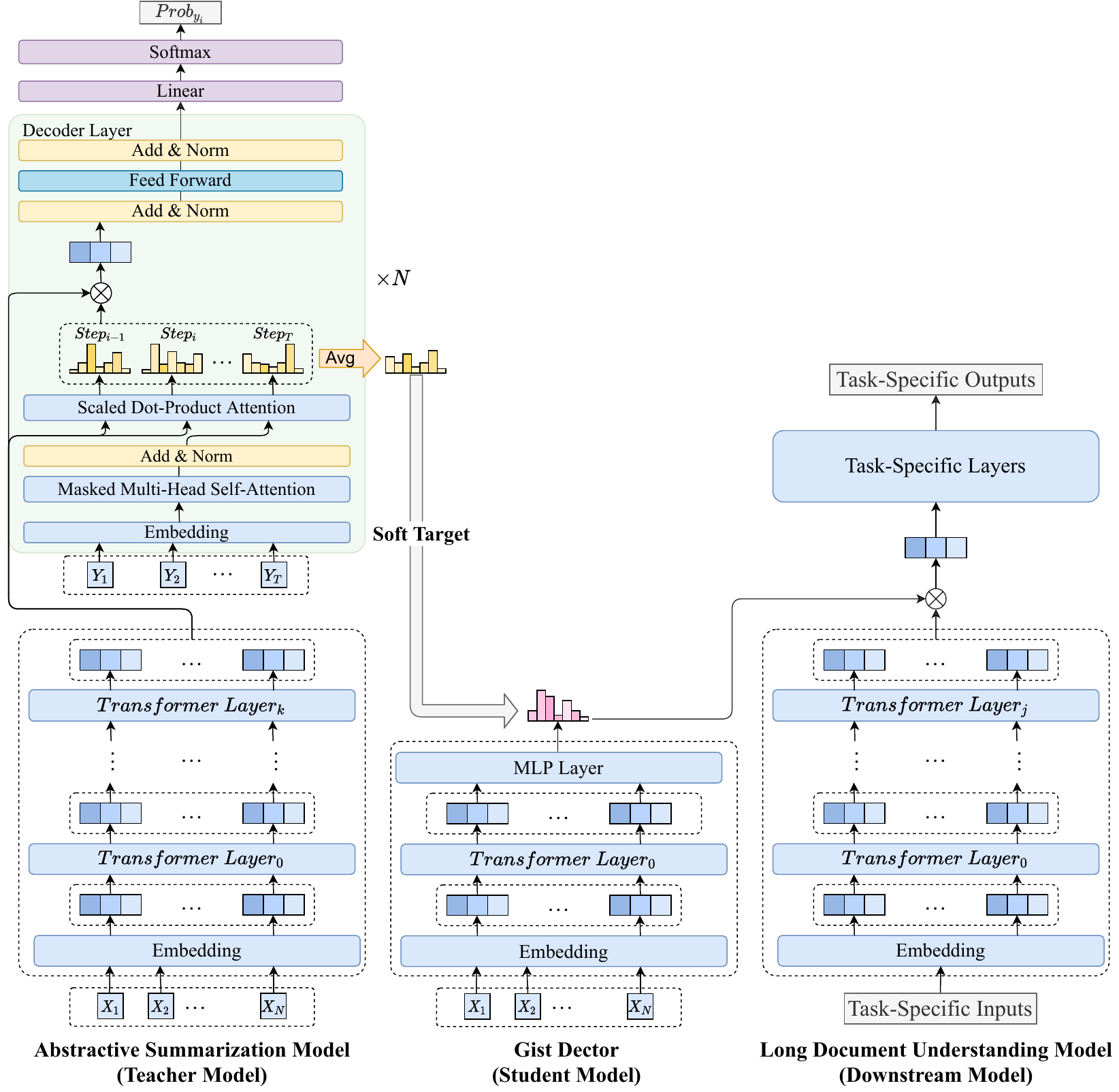}
		\caption{\label{fig2} Gist Detector is trained to reproduce the salient information from the teacher model, and provides the salient-aware representations as supplementary to augment the downstream model.}
	\end{center}
\vskip -0.2in
\end{figure}
\raggedbottom

\section{Methodology}
\label{sec:pagestyle}
In this paper, we propose our Gist Detector to leverage the gist detection ability of summarization model, and transfer gist information into downstream long text understanding models.
We first introduce the architecture of Gist Detector ( \S~\ref{ssec:architecture}).
During training, we use the knowledge distillation mechanism to transfer the gist detection ability from a well-trained summarization model (teacher model) to Gist Detector (student model) (\S~\ref{ssec:train}).
Then, we integrate gist information extracted by Gist Detector into downstream models (\S~\ref{ssec:inference}).
The much smaller model size and the non-autogressive architecture reduce the time-consuming problem, and the generated single gist-aware representation overcomes the mismatch problem.

\subsection{Gist Detector Architecture}
\label{ssec:architecture}
As shown in middle part of Figure~\ref{fig2}, Gist Detector has an encoder architecture, which learns the importance weight of each word in the source sequence from the summarization model, and produces this information for downstream models.
There are many possible network architectures for Gist Detector. 
We implement our Gist Detector with several Transformer encoder layers \cite{vaswani2017attention}, and show that the a simple distilled Gist Detector can successfully benefit the long document understanding models.

Specifically, the input $\{ x_1, ..., x_N \}$ is firstly mapped into embeddings $\{ \mathbf{e}_1, ..., \mathbf{e}_N \}$, then fed into a four-layer transformer encoder and obtain the representations $\mathbf{H} = \{ \mathbf{h}_1, ..., \mathbf{h}_N \}$.
Then, a two-layer MLP followed by a softmax function is applied to produce the the probability distribution over the input text $\mathbf{p} = \{ p_1, ..., p_N \}$, which reveals the importance of each word in the source sequence.

\subsection{Training with knowledge distillation}
\label{ssec:train}
We leverage the knowledge distillation mechanism to train Gist Detector (student model) with the salient information extracted from the abstractive summarization model (teacher model).
Different from the typical knowledge distillation, which uses the teacher's predictive distribution over the target classes as the soft target, we assume the attention distribution extracted from the decoding process reveals the salient information of the source text, and use the teacher's attention distribution as the soft target. The student model learns to reproduce the attention distribution for each training sample.

Specifically, the soft target $\mathbf{q} =\{ q_1, ..., q_N \}_{n=1}$ is calculated as the geometric mean of the attention distribution over all decoding steps:
\begin{equation}
	q_n = \frac{\sum_t a_{n, t}}{T}
\end{equation}
, where $T$ is the total decoding steps. 
Finally, the optimization objective is the cross entropy between the predicted probability distribution $\mathbf{p}$ of the student model and the soft target $\mathbf{q}$ from the teacher model: 
\begin{equation}
	L_{KD} = - \sum_{(x,y)} \sum_{n=1}^{N} q_n log(p_n))
\end{equation}

\begin{table*}
	\centering
 \vskip -0.4in
    \resizebox{1.0\textwidth}{!}{
        \begin{tabular}{l c c c c c cccccccccccc}
		\hline 
\textbf{Methods}	&	\textbf{Appeal}		&		\textbf{Baby}		&		\textbf{Books}		&		\textbf{Camera}		&		\textbf{DVD}			&		\textbf{Electronics}	&		\textbf{Health}		&		\textbf{IMDB}		&		\textbf{Kitchen}		&		\textbf{Magazines}	&		\textbf{MR}			&		\textbf{Music}		&		\textbf{Software}	&		\textbf{Sports}		&		\textbf{Toys}		&		\textbf{Video}	& \textbf{Overall}	\\
		\hline
ASP-MTL &	$87.0$	 &	$88.2$	 &	$84.0$	 &	$89.2$ &	$85.5$ &	$86.8$ &	$88.2$ &	$85.5$ &	$86.2$ &	$92.2$ &	$76.7$ &	$82.5$ &	$87.2$ &	$85.7$ &	$88.0$ &	$84.5$ &	$86.1$ \\
S-LSTM &	$85.8$ &	$86.3$ &	$83.4$ &	$90.0$ &	$85.5$ &	$83.3$ &	$86.5$ &	$87.2$ &	$84.5$ &	$93.8$ &	$76.2$ &	$82.0$ &	$87.8$ &	$85.8$ &	$85.3$ &	$86.8$ &	$85.6$ \\
Meta-MTL &	$87.0$ &	$88.$0 &		\textbf{$87.5$} &	$89.7$ &		\textbf{$88.0$} &	$89.5 $&		\textbf{$90.3$} &	$88.0$ &		\textbf{$91.3$} &	$91.0$ &	$77.0$ &	$86.3$ &	$88.5$ &	$86.7$ &		\textbf{$88.5$} &	$88.3$ &	$87.9$ \\
BiLSTM &	$84.8$ &	$84.5$ &	$78.8$ &	$86.3$ &	$80.7$ &	$81.9$ &	$83.0$ &	$79.8$ &	$82.1$ &	$90.5$ &	$76.1$ &	$79.6$ &	$85.4$ &	$80.$3 &	$83.9$ &	$81.1$ &	$82.4$ \\
		\hline
\rowcolor{LightGray}BiLSTM+GD &		\textbf{$87.6$}		&		\textbf{$88.5$}		&	$86.7$		&		\textbf{$90.8$}	&	$87.8$		&		\textbf{$89.6$}		&	$88.2$		&		\textbf{$88.1$}	&	$90.7$		&		\textbf{$94.6$}	&		\textbf{$78.2$}		&		\textbf{$86.4$}		&		\textbf{$90.3$}	&		\textbf{$87.1$}		&	$88.3$		&		\textbf{$88.5$}		&	\textbf{$88.2$}	\\
~~- w/o KD &	$86.8$ &	$86.4$ &	$81.7$ &	$89.1$ &	$82.9$ &	$82.3$ &	$84.5$ &	$81.2$ &	$85.0$ &	$91.7$ &	$76.5$ &	$82.9$ &	$87.8$ &	$84.9$ &	$85.4$ &	$84.6$ &	$84.6$\\

		\hline
	\end{tabular}}
\caption{
Document classification results across 16 domains of  FDU-MTL datasets.
GD denotes our Gist Detector method. KD denotes the knowledge distillation training.
}
\label{table_doc_classify}  
\vskip -0.1in
\end{table*}
\raggedbottom

\subsection{Integration of salient information}
\label{ssec:inference}
To enhance the long document understanding ability of the downstream model, we extract the salient information from the well-trained Gist Detector, and integrate it into the downstream model with a fuse module.

Specifically, for each long text $\{ x_1, ..., x_N \}$ as the input, the Gist Detector produces the probability distribution  $\mathbf{p} = \{p_1, ..., p_N \}$ over the input text, revealing the importance weights of each word. 
Given the context representation of the long document understanding model $\mathbf{c} = \sum_n^N \mathbf{s_n}$, we fuse the context representation $\mathbf{c}$ with the importance weights $\mathbf{p}$ as:
\begin{equation}
	\mathbf{c}' = (1 - \lambda) \mathbf{c} + \lambda \sum_t p_t\mathbf{s}_t
	\label{integrate_type_1}
\end{equation}
, where $\lambda \in [0,1]$ is a tunable hyperparameter.
As for the downstream model that predict scores for each word of the input text, such as extractive QA models, we fuse the prediction scores $\{r_1, ..., r_N\}$ with the importance weights $\{p_1, ...,  p_N \}$: 
\begin{equation}
	r_{t}' = (1 - \lambda') r_{t} + \lambda p_t
	\label{integrate_type_2}
\end{equation}
Note that we use the importance weight rather than the context representation as the salient information, since it contains much less parameters and alleviates the impact of domain-specific information.

\section{Experiments}

\subsection{Distillation}
\label{ssec:appendix_training}
Firstly, We train an ensemble of $8$ abstractive summarization models with Transformer-based encoder-decoder architecture as the teacher model on \emph{CNN/Daily Mail}. The average ROUGE ${\rm F}_{1}$ scores \cite{lin2004rouge} of the teacher model are $38.6$, $16.3$ and $35.4$ for ROUGE-1, ROUGE-2 and ROUGE-L respectively.
We follow the same setup and use the scripts provided by \cite{see2017get} to pre-process the \emph{CNN/Daily Mail} dataset. 
We use the $100$ dimensional filters with width of $5$ for CNN to capture the character embeddings. 
We select the $300$d GloVe pre-trained word embedding and share the same word embedding weight between encoder and decoder. 
The hidden size of Transformer is $512$. We use the Adam optimizer \cite{kingma2014adam} with learning rate of $0.0004$, $\beta_1$ = $0.9$, $\beta_2$ = $0.999$. The dropout rate and batch size are set to $0.35$ and $16$, respectively. To avoid the gradient explosion problem, we apply the gradient norm clipping with a maximum gradient norm of $2.0$. 

Then we train Gist Detector with Transformer-based encoder architecture using knowledge distillation mechanism. We use 100d GloVe for word embedding, $50$d for character embedding, the hidden size for the Transformer encoder is $256$. We take the same optimization setting as that of the teacher model.

\subsection{Integration into Downstream Tasks}
\label{sec:appendix_integration}
Finally, we transfer the salient information from the well-trained Gist Detector to downstream models of three long text understanding tasks: document classification, distantly supervised open-domain question answering (DS-QA) and non-parallel text style transfer.

\subsubsection{Document Classification: }
We take the BiLSTM model as our baseline model for document classification task that concatenates the final state values of forward and backward pass as the context representation vector, then feeds it into a MLP to predict the label.
We initialize the word embedding with the $300$d GloVe.
The hidden size of BiLSTM is set as $256$. The layer number of BiLSTM and MLP are both set to $2$. We take the Adam as optimizer with lr = $0.001$, $\beta_1$ = $0.9$, $\beta_2$ = $0.999$, $0.35$ dropout and train for $6$ epochs.
The $\lambda$ in \S~\ref{ssec:inference} is set to be $0.5$ while integrating the BiLSTM model with our Gist Detector.

\subsubsection{Distantly Supervised Open-Domain QA:}
We use the OpenQA model\cite{lin2018denoising} as our baseline model for distantly supervised open-domain question answering task, which applies a selector to filter passages, then a precise reader to extract the potential answers, finally aggregates these results to predict the final answer. 
We evaluate our method on two high-quality datasets, \emph{TriviaQA} (open-domain setting)\cite{joshi2017triviaqa} and \emph{SearchQA}\cite{dunn2017searchqa} with two metrics including ExactMatch (EM) and F1 scores.
We keep the same setup of hyper-parameters and training settings as that in OpenQA while some important details are as follow. We combine the passage selector with Gist Detector as introduced in \S~\ref{ssec:inference} and the $\lambda$ is set as 0.5. We feed the $\mathbf{c}'$ through a linear function followed by multiplication with the question vector to produce the score for filtering passages and add it to the original score produced by the OpenQA selector to predict the final passage score. For the reader, we directly add the predicted score of answer span with the probability distribution $\mathbf{p}$ produced by Gist Detector as introduced in \S~\ref{ssec:inference} to produce the final score, where the $\lambda'$ is set as $0.2$.

\subsubsection{Text Style Transfer:}
As for the non-parallel text style transfer task, the model aims to compress gist of texts into fixed-size vectors separated from pure style information. 
We select Cross-aligned AE\cite{shen2017style} and Adversarially Regularized Autoencoder (ARAE)\cite{kim2017adversarially} as our baseline models.
We follow the setup of \cite{shen2017style} but remain reviews whose length are between $70$ and $150$ rather than not exceeding $15$, and eventually obtain $350$K, $280$K non-parallel data from Amazon and Yelp reviews respectively. We keep the same setup of hyper-parameters and training settings as that of Cross-aligned AE and ARAE. We combine the content vector with our Gist Detector as introduced in \S~\ref{ssec:inference}, and the $\lambda$ is set to be $0.5$.
To evaluate the model, we use $4$ automatic metrics: (i) Acc: the accuracy of successfully changing the style into the target style measured by a pre-trained classifier. Following \cite{shen2017style}, we use the TextCNN model as the classifier that achieves the accuracy of $94.2\%$ and $95.7\%$ on Amazon and Yelp respectively. (ii) Cosine: we follow the setup of \cite{fu2018style} to measure the content preservation with cosine similarity. 
(iii) Entity: we use the proportion of noun entities to measure the content consistency between source and generated texts. (iv) PPL: the fluency of generated texts measured by a pre-trained language model on corresponding datasets.

\section{Results and Analysis}
\label{sec:results}

\subsection{Results on Document Classification}
\label{ssec:architecture}
We evaluate our our method across $16$ domains on FDU-MTL datasets\cite{liu2017adversarial}.
As shown in Table \ref{table_doc_classify}, augmented with our Gist Detector, the baseline BiLSTM model obtains significant performance improvement on all of the $16$ domains and outperforms prior approaches (ASP-MTL \cite{liu2017adversarial}, S-LSTM \cite{zhang2018sentence}, Meta-MTL \cite{chen2018meta}) with $88.2$ overall accuracy.
An ablation study shows that if we use Gist Detector with random initializad parameters, the overall performance drops $3.6$. It indicates that both the additional parameters from Gist Detector and the gist detection ability distilled from the summarization model contributes to the performance improvement.

\begin{table}[tb]
\centering
\small
\footnotesize
\vskip -0.3in
\resizebox{0.35\textwidth}{!}{
\begin{tabular}{l c c c c c}
\hline
\multirow{2}*{\textbf{QA Models}} 
& \multicolumn{2}{c}{\textbf{TriviaQA}} 
&  \multicolumn{2}{c}{\textbf{SearchQA}} \\ \cline{2-5} 
& \textbf{EM} & \textbf{F1}    & \textbf{EM}     & \textbf{F1}      \\ \hline
BiDAF \cite{seo2016bidirectional}	&	- &	- &	$28.6$ &	$34.6$		\\
AQA \cite{buck2018ask}			&	- &	- & $40.5$	 &	$47.4$		\\
R$^{3}$ \cite{wang2018r3}	 & $47.3$ & $53.7$ & $49.0$ & $55.3$ \\
Re-Ranker \cite{wang2018evidence}	&	\textbf{$50.6$} &	$57.3$ & $57.0$	 &	$63.2$	\\
TraCRNet \cite{dehghani2019learning} & - & - & $52.9$	 &	\textbf{$65.1$} \\
OpenQA \cite{lin2018denoising} & $48.7$ & $56.3$  & $58.8$  & $64.5$ \\ 
\hline
\rowcolor{LightGray}OpenQA + GD &  $50.3$ & \textbf{$57.6$} & \textbf{$59.5$} &  \textbf{$65.1$}   \\
~~- w/o GD in selector    &  $49.2$ & $56.5$ & $59.0$ &  $64.8$     \\
~~- w/o GD in reader  &  $49.4$ & $57.1$ & $59.2$ &  $64.8$     \\
\hline
\end{tabular}}
\caption{EM and F1 results on the TriviaQA (open-domain setting) and SearchQA datasets. 
}
\label{table_qa_performance}
\end{table}
\raggedbottom

\begin{table}[tb]
	\centering
    \resizebox{0.45\textwidth}{!}{
	\begin{tabular}{l c c c c c c c}
		\hline
		\multirow{2}*{\textbf{QA Models}} 
			& \multicolumn{3}{c}{\textbf{TriviaQA}} 
			&  \multicolumn{3}{c}{\textbf{SearchQA}} \\ \cline{2-7} 
		& \textbf{Hit@1} & \textbf{Hit@3}    & \textbf{Hit@5} 
		& \textbf{Hit@1} & \textbf{Hit@3}    & \textbf{Hit@5}      \\ \hline
		OpenQA		&	$43.4$ &	$51.5$ &	$54.5$ &	$59.1$ &	$68.7$ &	$76.3$	\\
		\rowcolor{LightGray}OpenQA + GD	&	$49.1$ &	$57.7$ &	$63.1$ &	$65.3$ &	$73.4$ &	$79.6$	\\
		\hline
	\end{tabular}}
	\caption{Performance of passage selection on TriviaQA and SearchQA development set. Hit@N represents the proportion of related passages being ranked in top-N.}
	\label{table_qa_hit}  
 \vskip -0.1in
\end{table}
\raggedbottom

\subsection{Results on DS-QA}
We evaluate our method on TriviaQA (open-domain setting) \cite{joshi2017triviaqa} and SearchQA \cite{dunn2017searchqa} datasets with ExactMatch (EM) and F1 score metrics. 
As shown in Table \ref{table_qa_performance},
Augumented with our Gist Detector, the baseline OpenQA model performs much better on both two datasets.
An ablation study shows that integration of salient information into both the selector and the reader leads to the best performance.
Table \ref{table_qa_hit} shows the passage selection performance of our method.
We find that with Gist Detector, the selector filters passages much more precisely, thus our QA system can aggregate information among fewer passages and make faster answer predictions.

\begin{table}[tb]
\centering
\vskip -0.3in
\resizebox{0.45\textwidth}{!}{
\begin{tabular}{c c c c c c c c c}
\hline
\multirow{2}*{\textbf{Models}} 
& \multicolumn{4}{c}{\textbf{Amazon}} 
&  \multicolumn{4}{c}{\textbf{Yelp}} \\ \cline{2-9} 
& \textbf{Acc} & \textbf{Cosine} & \textbf{Entity} & \textbf{PPL} 
& \textbf{Acc} & \textbf{Cosine} & \textbf{Entity} & \textbf{PPL}	\\ \hline
Cross-aligned AE	 &	$84.7\%$ &	$0.46$ &	$26.13$ &	$34.67$ &	$89.5\%$ &	$0.53$ &	$26.63$ &	$28.46$	\\
ARAE			 &	$86.2\%$ &	$0.57$ &	$31.37$ &	$36.36$ &	$89.3\%$ &	$0.61$ &	$32.46$ &	$29.18$	\\
\rowcolor{LightGray}ARAE + GD		 &	$91.0\%$ &	$0.71$ &	$47.56$ &	$24.15$ &	$93.4\%$ &	$0.73$ &	$49.04$ &	$21.43$	\\
\hline
\end{tabular}}
\caption{
Automatic evaluation results on Amazon and Yelp text style transfer datasets. 
}
\label{table_tst_performance}  
\end{table}
\raggedbottom

\subsection{Results on Text Style Transfer}
We further evaluate our method on the Amazon and Yelp text style transfer dataset \cite{shen2017style}.
The automatic evaluation results from Table \ref{table_tst_performance} shows that with our Gist Detector, the baseline model ARAE\cite{kim2017adversarially} can achieve significantly higher transfer accuracy, better content preservation, better noun entity preservation and much more fluency. 
It indicates that the Gist Detector helps the model detect and compress more important information from long texts. 
Moreover, we conduct human evaluation to further evaluate the quality of the style transfer models.
We randomly select $1000$ examples ($500/500$ positive/negative), and employ people to judge whether texts are converted to the target style, and to evaluate content correlation ($0-5$, $5$ for the most correlative) and fluency ($0-5$, $5$ for the most fluent). 
As shown in Table \ref{table_tst_human_evaluation}, Gist Detector can significantly improve the baseline model’s performance on all evaluation metrics.

\begin{table}[tb]
\centering
\resizebox{0.5\textwidth}{!}{
\begin{tabular}{c c c c c c c}
\hline
\multirow{2}*{\textbf{Models}} 
& \multicolumn{3}{c}{\textbf{Amazon}} 
&  \multicolumn{3}{c}{\textbf{Yelp}} \\ \cline{2-7} 
& \textbf{Acc} & \textbf{Correlation} & \textbf{Fluency} 
& \textbf{Acc} & \textbf{Correlation} & \textbf{Fluency}	\\ \hline
Cross-aligned AE	 & 56.4\% & 2.4 & 3.0 & 58.2\% & 2.7 & 3.1	 \\
ARAE			 & 73.6\% & 2.8 & 3.3 & 74.1\% & 3.1 & 3.5	 \\
\rowcolor{LightGray}ARAE + GD		 & 78.2\% & 3.7 & 3.5 & 78.6\% & 3.9 & 3.8 \\
\hline
\end{tabular}}
\caption{Human evaluation on accuracy, content correlation and fluency of the generated text.}
\label{table_tst_human_evaluation}  
 \vskip -0.1in
\end{table}

\section{Conclusion}
\label{sec:print}
In this paper, we propose Gist Detector to learn gist detection ability from a summarization model with knowledge distillation mechanism. We integrate the gist information detected by distilled Gist Detector into different downstream models to enhance their long document understanding ability.
Experimental results show that our method significantly improves the performance of all baseline models for different tasks that require long text understanding. 
Future work will involve finding better strategies to integrate our gist detector into more tasks and processing longer sequences.

\vfill\pagebreak

\bibliographystyle{IEEEbib}
\bibliography{refs}

\end{document}